\documentclass{article}

\usepackage[preprint]{neurips_2023}

\usepackage[utf8]{inputenc} 
\usepackage[T1]{fontenc}    
\usepackage{hyperref}       
\usepackage{url}            
\usepackage{booktabs}       
\usepackage{amsfonts}       
\usepackage{nicefrac}       
\usepackage{microtype}      
\usepackage{xcolor}         

\usepackage{tabularx} 
\usepackage{ragged2e} 
\usepackage{booktabs}
\usepackage{makecell} 

\usepackage{graphicx}
\usepackage{booktabs}
\usepackage{multirow}
\usepackage{colortbl}
\usepackage{amsmath}
\usepackage[capitalize]{cleveref}
\usepackage{natbib}
\usepackage{enumitem} 

\definecolor{Gray}{gray}{0.9}
\definecolor{Keyword_red}{HTML}{EFB8B4}
\definecolor{Sentence_green}{HTML}{C1DCC4}

\definecolor{Paragraph_blue}{HTML}{B7CBE0}

\setcitestyle{numbers}
\setcitestyle{square}
\setcitestyle{comma}
\usepackage{graphicx}
\usepackage{subcaption} 
\usepackage{latexsym}

\newcommand{\modelname}{\textsc{QwenLong-CPRS}}

\title{\modelname: Towards $\infty$-LLMs with Dynamic Context Optimization}

\author{Weizhou Shen, Chenliang Li, Fanqi Wan, Shengyi Liao, Shaopeng Lai, Bo Zhang, \\ \bf{Yingcheng Shi, Yuning Wu, Gang Fu, Zhansheng Li, Bin Yang,} \\
\bf{Ji Zhang, Fei Huang, Jingren Zhou, Ming Yan\thanks{$\;$Corresponding author}}\\
Tongyi Lab, Alibaba Group\\
}

\def\huggingface{\raisebox{-1.5pt}{\includegraphics[height=1.05em]{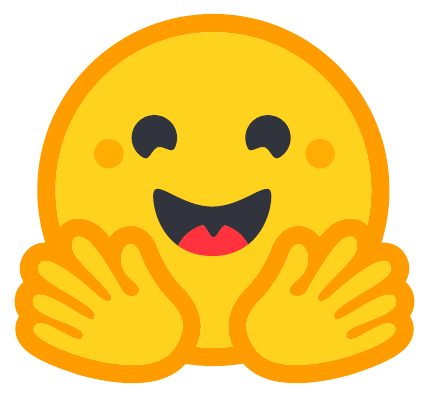}}}
\def\modelscope{\raisebox{-1.5pt}{\includegraphics[height=0.85em]{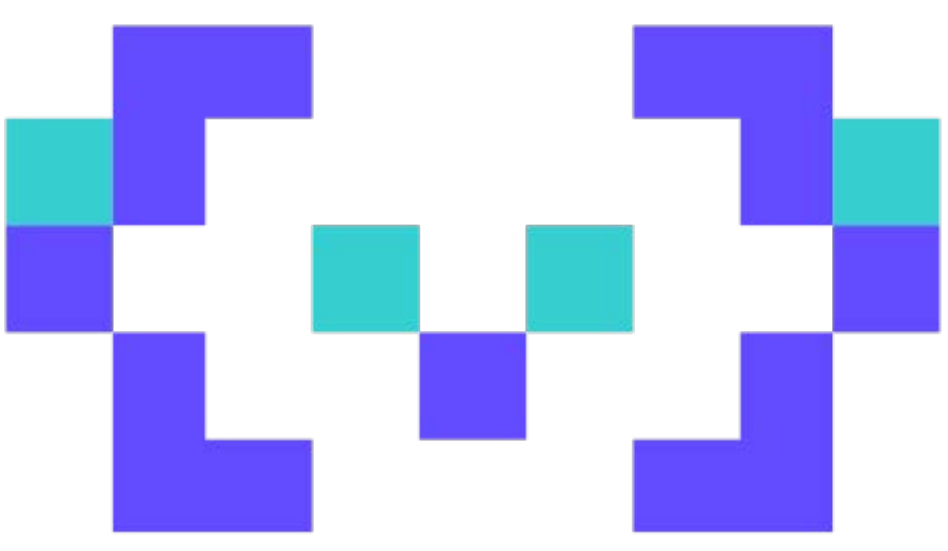}}}
\def\github{\raisebox{-1.5pt}{\includegraphics[height=1.05em]{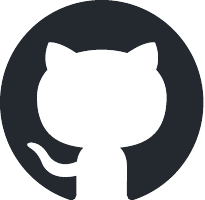}}}

\newcommand{\hflink}{https://huggingface.co/Tongyi-Zhiwen/QwenLong-CPRS-7B}
\newcommand{\mslink}{https://modelscope.cn/models/iic/QwenLong-CPRS-7B}
\newcommand{\ghlink}{https://github.com/Tongyi-Zhiwen/QwenLong-CPRS}

\begin{document}

\maketitle

\begin{center}
\vspace{-0.8cm}
\begin{tabular}{rc}
\github & \url{\ghlink}\\
\huggingface & \url{\hflink}\\
\modelscope & \url{\mslink}\\
\end{tabular}
\vspace{0.2cm}
\end{center}


\begin{abstract}
    This technical report presents {\modelname}, a context compression framework designed for explicit long-context optimization, addressing prohibitive computation overhead during the prefill stage and the ``lost in the middle'' performance degradation of large language models (LLMs) during long sequence processing. Implemented through a novel \emph{dynamic context optimization} mechanism, {\modelname} enables multi-granularity context compression guided by natural language instructions, achieving both efficiency gains and improved performance.

    Evolved from the Qwen architecture series, {\modelname} introduces four key innovations: (1) Natural language-guided dynamic optimization, (2) Bidirectional reasoning layers for enhanced boundary awareness, (3) Token critic mechanisms with language modeling heads, and (4) Window-parallel inference. 


    Comprehensive evaluations across five benchmarks (4K-2M word contexts) demonstrate {\modelname}'s threefold effectiveness: (1) Consistent superiority over other context management methods like RAG and sparse attention in both accuracy and efficiency. (2) Architecture-agnostic integration with \emph{all} flagship LLMs, including GPT-4o, Gemini2.0-pro, Claude3.7-sonnet, DeepSeek-v3, and Qwen2.5-max, achieves 21.59$\times$ context compression alongside 19.15-point average performance gains; (3) Deployed with Qwen2.5-32B-Instruct, {\modelname} surpasses leading proprietary LLMs by 4.85 and 10.88 points on Ruler-128K and InfiniteBench, establishing new SOTA performance. 
\end{abstract}

\begin{figure*}[h]
    \centering

    \begin{subfigure}[b]{0.52\linewidth} 
        \centering
        \includegraphics[width=1.0\linewidth]{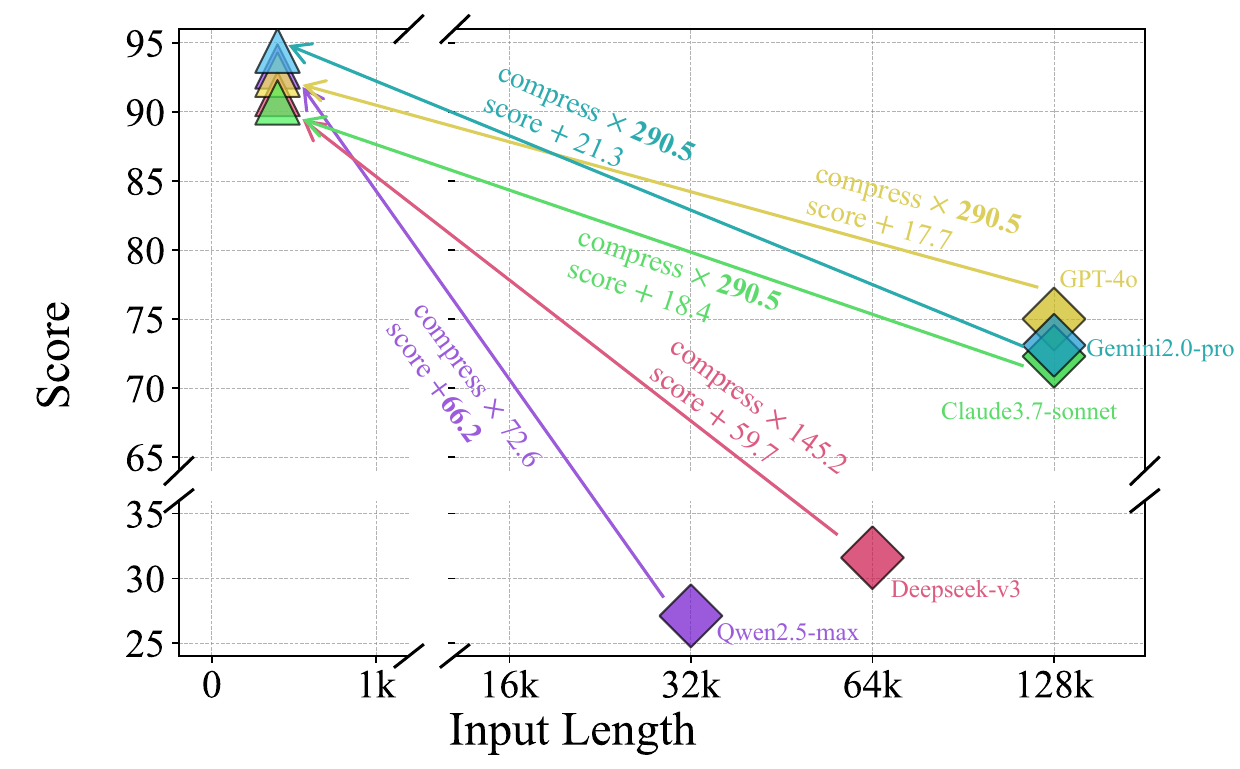}
        \vspace{-0.3cm}
        \caption{The input compression rate and performance gain when different LLMs are cascaded with {\modelname}.}
        \label{fig:intro_compress_and_performance}
    \end{subfigure}
    \hfill 
    \begin{subfigure}[b]{0.45\linewidth} 
        \centering
        \includegraphics[width=1.0\linewidth]{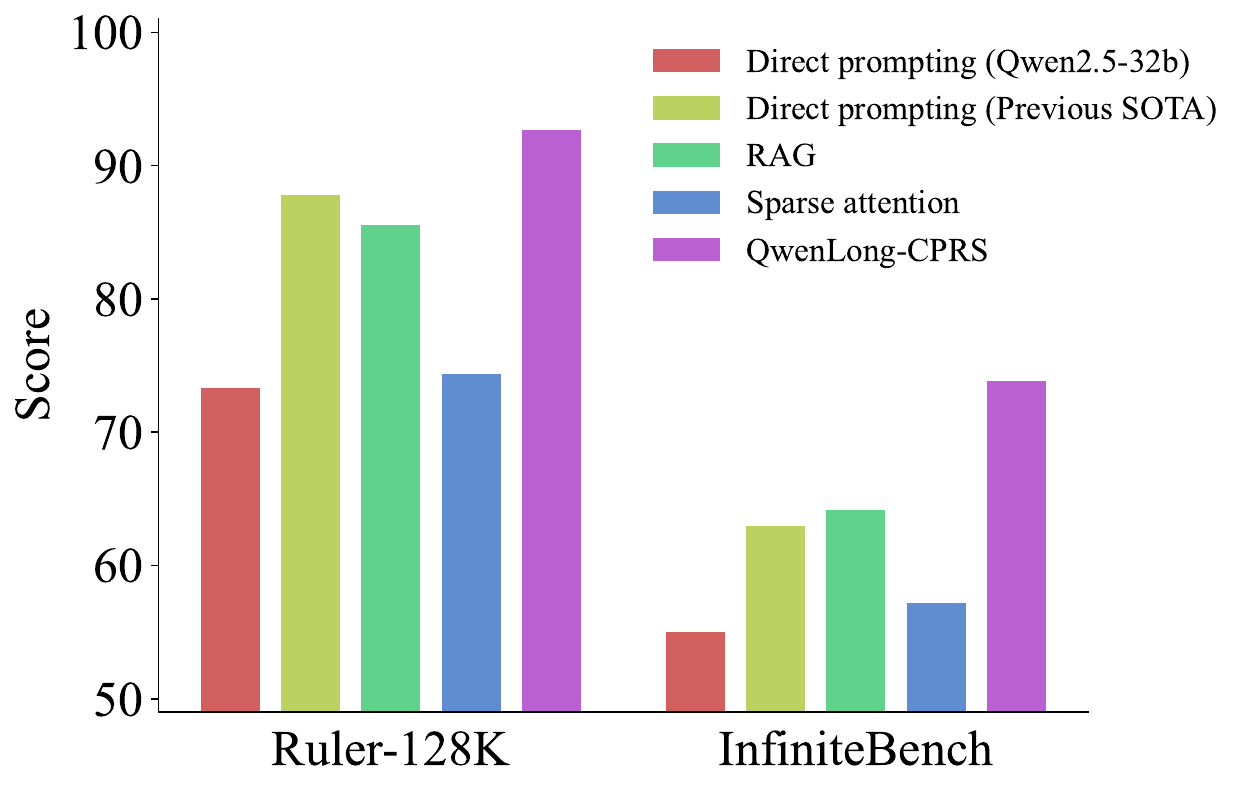}
        \vspace{-0.3cm}
        \caption{Performance of Qwen2.5-32b-instruct with different context management methods.}
        \label{fig:intro_method_compare}
    \end{subfigure}

    \caption{Illustration of the performance of {\modelname}.  Figure ~\ref{fig:intro_compress_and_performance} compares the input token consumption and model performance of various LLMs on Ruler-128K before (marked with $\Diamond$) and after (marked with $\Delta$) cascading {\modelname}. Figure \ref{fig:intro_method_compare} highlights the performance improvements of {\modelname} over other context management methods, such as RAG~\cite{gteembedding} and sparse attention~\cite{jiang2024minference}.
 }
    \label{fig:combined}
\end{figure*}


\section{Introduction}

Enhancing the long-context processing capabilities of large language models (LLMs) has emerged as a critical research frontier in both academia and industry~\cite{openai2024gpt4technicalreport,geminiteam2024gemini15unlockingmultimodal,qwen2025qwen25technicalreport,grattafiori2024llama3herdmodels}. Recent advancements in positional embedding techniques~\cite{su2023roformerenhancedtransformerrotary,peng2023yarnefficientcontextwindow} and the curation of synthetic long-context training data~\cite{bai2024longalignrecipelongcontext} have enabled significant progress in extending LLMs' context windows, achieving expansions from 4K to over 1M tokens~\cite{yang2025qwen251mtechnicalreport, gemini25}. Despite these advances, two critical challenges persist. First, the quadratic computational complexity of processing long sequences imposes prohibitive efficiency costs. Second, the unresolved ``lost in the middle'' phenomenon~\cite{liu2023lostmiddlelanguagemodels}, where LLMs struggle to effectively prioritize critical information within lengthy inputs.


A fundamental strategy involves efficiently managing long context by focusing on key content within the model's reliable context window~\cite{liu2025spakelongcontextlargelanguage}. Building upon this strategy, two primary approaches have been proposed: Retrieval-augmented generation (RAG) frameworks~\cite{lewis2021retrievalaugmentedgenerationknowledgeintensivenlp,chen2023walkingmemorymazecontext} enhance computational efficiency by dynamically retrieving query-relevant text chunks from input contexts, enabling selective processing of contextual information. Conversely, sparse attention (SA) mechanisms~\cite{jiang2024minference,MOBA,NSA} redesign the self-attention mechanism within LLMs, either by restricting attention computations to structured patterns or by prioritizing critical token interactions during sequential generation. 

Despite their advantages, both approaches exhibit significant limitations. First, RAG systems, while efficient, rely on coarse-grained chunk-level embeddings, leading to imprecise outputs. This limitation becomes particularly problematic in scenarios requiring fine-grained localization of uniformly distributed knowledge~\cite{li2024retrievalaugmentedgenerationlongcontext}. On the other hand, SA methods, though flexible in token-level aggregation, necessitate substential data construction and computationally intensive model training to optimize attention patterns~\cite{MOBA,NSA}, alongside specialized infrastructure investments.

\begin{figure*}[t]
    \centering
    \includegraphics[width=1.0\linewidth]{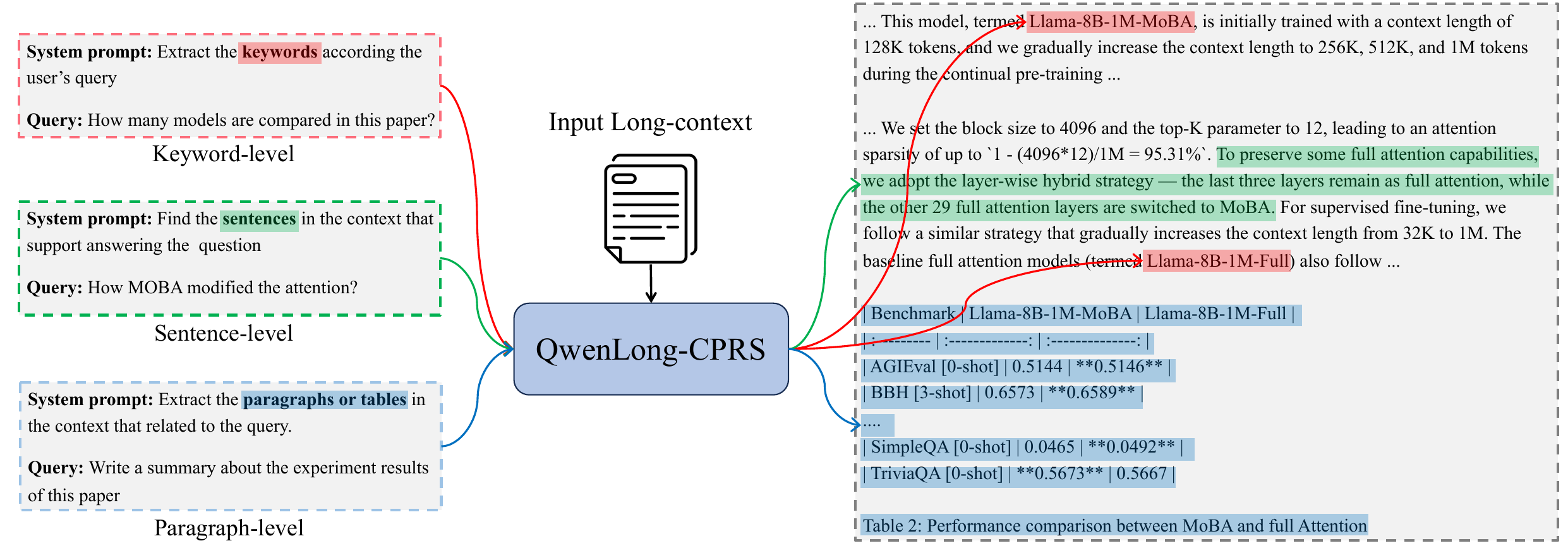}
    \caption{The concept of \textit{dynamic context optimization}, which aims to enhance context processing efficiency by maximizing information density. Given a long-context input, this paradigm dynamically compresses it into query-specific content at varying granularities, facilitating concise and accurate information extraction for different user queries. For instance, \textcolor{Keyword_red}{keywords} for \textit{search queries}, \textcolor{Sentence_green}{sentences} for \textit{question answering}, and \textcolor{Paragraph_blue}{paragraphs} for \textit{summarization}.} 
    \label{fig:concept}
    \vspace{-0.3cm}
\end{figure*}

To address these challenges, we introduce an innovative \textit{dynamic context optimization} paradigm, which aims to improve context-processing efficiency through maximizing information density. As depicted in Figure \ref{fig:concept}, this approach dynamically compresses input contexts into query-tailored segments across different granularities, enabling concise and accurate context optimization for various user queries. This paradigm advances existing methods in two key aspects. First, it replaces RAG's coarse chunk-level retrieval with precise token-level content selection, enhancing information identification accuracy. Second, it operates independently as a plug-and-play component, eliminating SA's requirement for model retraining while maintaining compatibility with any downstream LLMs. 

Building upon the dynamic context optimization paradigm, we propose a novel compression system {\modelname}. Specifically, {\modelname} takes the control prompt, task query, and long context as input, and then labels the token critic score to compress task-relevant content based on a single forward pass. To endow {\modelname} with both precise and controllable characteristics, we redesign the attention mechanism into a hybrid architecture that combines bi-directional language modeling for comprehensive context location and causal language modeling for reliable language representation. Additionally, we develop a language modeling as token critic framework that repurposes the existing LLM's language modeling head to label token-level importance scores, thus maintaining the pretrained knowledge for better context compression

As illustrated in Figure~\ref{fig:intro_compress_and_performance}, {\modelname} demonstrates a remarkable context compression effect, achieving a context compression rate ranging from 72.6 to 290.5 times. This indicates that {\modelname} possesses efficient context optimization capabilities and can be seamlessly reused across various large models. More experimental results in Section~\ref{sec:expriment} across four long-context benchmarks whose input context lengths ranging from 4K to 2M tokens demonstrate {\modelname}'s superiority over direct prompting, RAG, and SA methods, with remarkable performance improvements and considerably higher inference efficiency. Notably, we show that smaller, short-context LLMs augmented with {\modelname} can outperform larger long-context counterparts. These findings highlight the potential of context optimization paradigms, offering a scalable and efficient pathway to augment LLMs' long-context processing capabilities.

Our key contributions are as follows:

\begin{itemize}[leftmargin=1em]

    \item We introduce dynamic context optimization, a novel paradigm for long-context management through dynamic, instruction-guided token-level compression. This paradigm optimizes information retention while adaptively prioritizing critical content.
    
    \item We propose {\modelname}, an innovative model that advances long-context processing via token-level critical scoring, enabling granular context optimization without sacrificing precision. We transformed the Qwen series models into an dynamic context optimization model by integrating a hybrid attention mechanism and leveraging the language modeling head token critic module. Furthermore, we systematically designed the framework for training data construction.

    \item Through extensive evaluation across four distinct benchmarks, we demonstrate that {\modelname} achieves substantial performance gains while reducing inference overhead. The framework shows consistent efficacy across LLMs of varying parameter scales and context lengths, establishing its versatility as a context-optimization-augmentation solution for long-context processing. 

\end{itemize}

\section{{\modelname}}\label{sec:method}

In this section, we initially present the formal definition of dynamic context optimization in Section~\ref{sec:formulation}. Subsequently, Section~\ref{sec:model_architecture} details the model architecture of {\modelname} and elaborates on the training methodology employed to align {\modelname} with the objectives of dynamic context optimization. Our effective window-parallelism inference method is also introduced in Section~\ref{sec:model_architecture}. Lastly, we describe the data construction method we have devised in Section~\ref{sec:data_construct}.

\subsection{Dynamic Context Optimization}\label{sec:formulation}

The input structure for long-context tasks comprises two components: the user query $q$, and the long context $X_{l}$. When $X_{l}$ exceeds the effective window size of LLMs, it typically causes either essential input truncation or the "lost-in-the-middle" phenomenon, ultimately degrading response quality. To address this, we propose identifying an information-condensed subset $X_s$ such that:
\begin{equation}
    |X_{s}|  \ll |X_{l}|, \quad \text{where} \quad X_{s}\subseteq X_{l}
\end{equation}

This process, termed \emph{context optimization}, aims to find the minimal-length 
$|X_s|$ that preserves maximally informative content for generating high-quality responses $Y$. Formally, we define our objective function as:
\begin{equation}
\mathcal{J} = \max_{\phi}\ \mathbb{E}_{X_s \subseteq X_l} \left[ \frac{\mathcal{I}(Y;[X_s,q])}{|X_s|^\beta} \right],
\end{equation}

where $\mathcal{I}(\cdot,\cdot)$ is the mutual information, $\beta$ controls the length penalty intensity, and $\phi$ parameterizes the context optimizer. In this paper, we propose {\modelname}, which achieves context optimization by  identifying and retaining the most semantically crucial tokens from $X_l$.  In addition, we introduce a natural language prompt $P$ that enables users to dynamically configure the granularity of the optimized context and how it will contribute the the response. Therefore, the resulting dynamically optimized context $X_s$ is formalized as:
\begin{equation}
    X_s = \mathcal{F}_{\phi}(P,q,X_l),
\end{equation}
where $\mathcal{F}(\cdot)$ is the token selection operation.

\subsection{Model Architecture}\label{sec:model_architecture}




\begin{figure*}[t]
    \centering

    \begin{subfigure}[b]{0.39\linewidth} 
        \centering
        \includegraphics[width=1.0\linewidth]{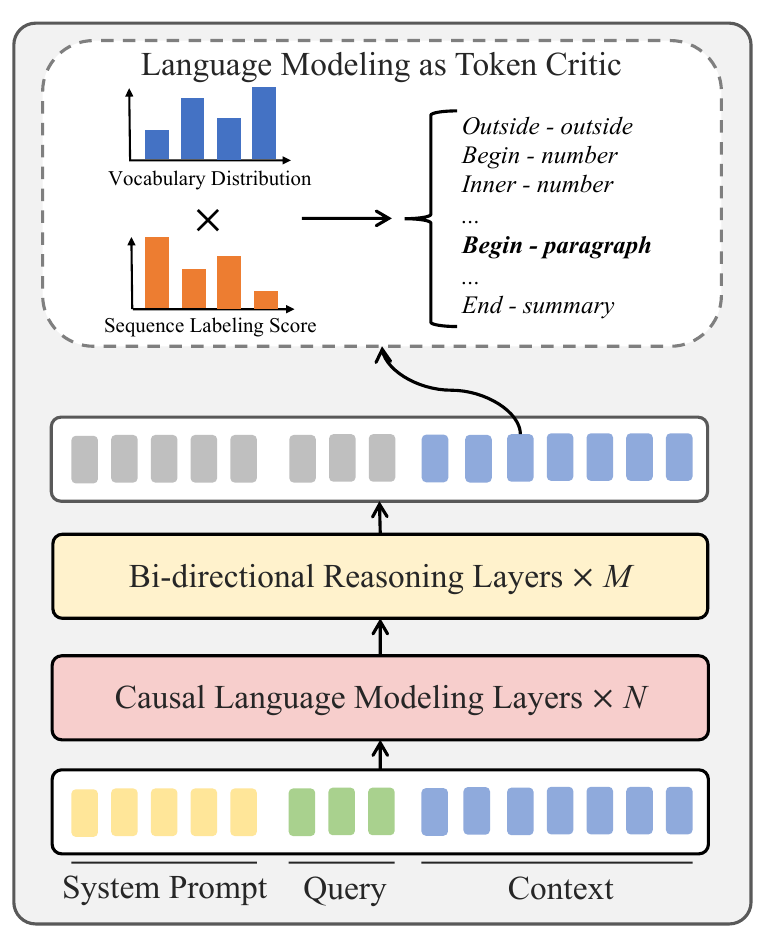}
        \vspace{-0.2cm}
        \caption{The model architecture of {\modelname}.}
        \label{fig:model_achitech}
    \end{subfigure}
    \hfill 
    \begin{subfigure}[b]{0.56\linewidth} 
        \centering
        \includegraphics[width=1.0\linewidth]{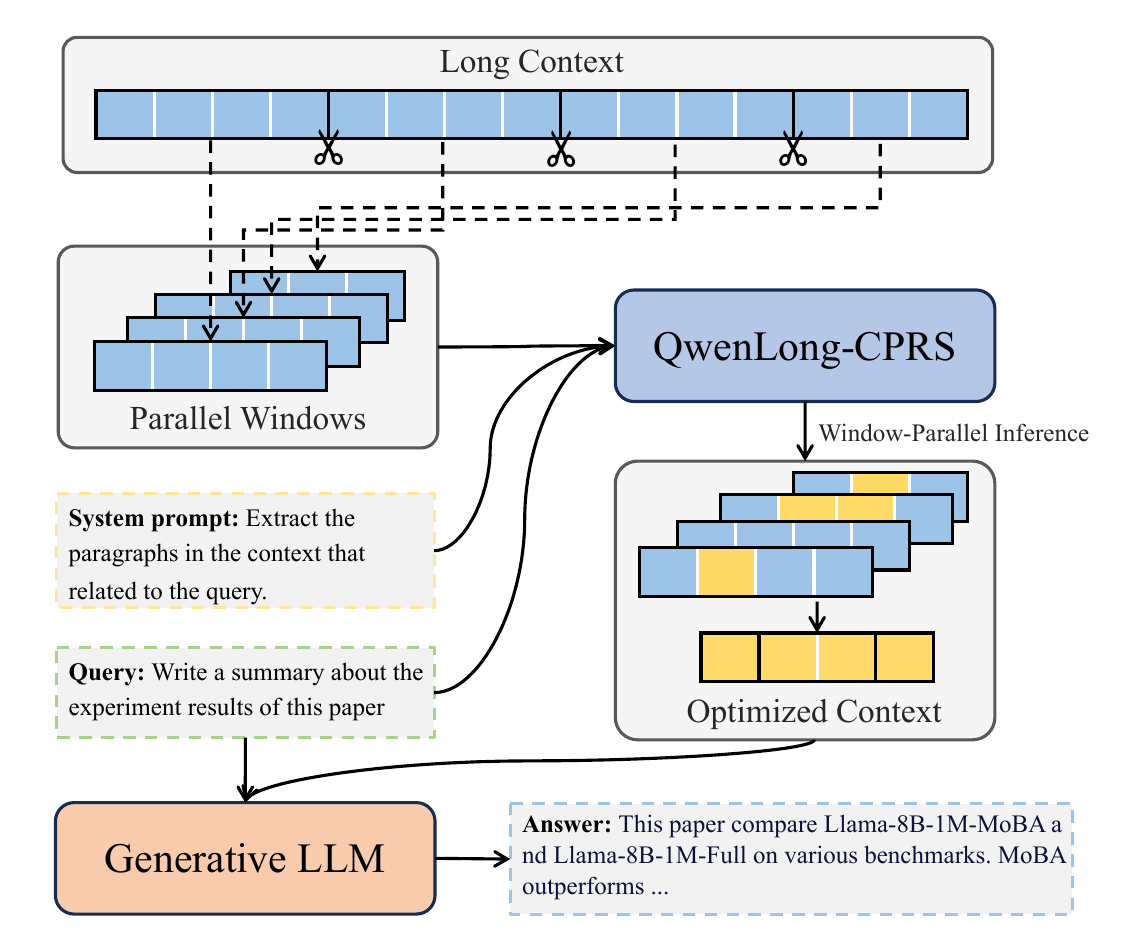}
        \vspace{-0.2cm}
        \caption{The workflow of generative LLMs cascading {\modelname} in this paper.}
        \label{fig:parrallel_compress}
    \end{subfigure}

    \caption{The model architecture and workflow of {\modelname}.}
    \label{fig:combine_framework}
    \vspace{-0.5cm}
\end{figure*}

The model architecture of {\modelname} is illustrated in Figure~\ref{fig:model_achitech}. To ensure model generalization and achieve better awareness of the use's controlling prompt and query, we adopt the parameters and vocabulary of the Qwen-2-Base~\cite{yang2024qwen2technicalreport} series models\footnote{Our experimental results demonstrate that the Qwen-2-Base series models outperform the Qwen-2.5-Base series models} as the starting checkpoint for initializing our {\modelname}. Beside the parameter initialization and vocabulary inheritance, there are several key modifications for {\modelname} to achieve better context optimization performance: 

\noindent\textbf{Dynamic Control via Natural Language: } We define the input structure for {\modelname} as a concatenation of three text components: \texttt{\{system prompt, user query, long context\}}, formatted according to the Qwen message organization template. In this paradigm, the system prompt specifies the desired properties of the optimized context, such as its granularity and its relationship to the query. For instance: \texttt{You are an expert in information extraction; your task is to extract sentences from the documents that support the user's question.} The user query contains the original instruction provided by the user, and the long context component is the source document requiring optimization. Through this paradigm design, {\modelname} can adaptively extract informative context according to the requirements of the system prompt and user query.


\noindent\textbf{Bidirectional Location Reasoning: } {\modelname} implements context optimization through a token scoring paradigm, preserving original token positions for prediction outputs. To improve boundary detection accuracy in optimized contexts, we introduce a \emph{bidirectional location reasoning} mechanism that enables global awareness of long-range contextual dependencies. As shown in Figure~\ref{fig:model_achitech},  we retain the causal masking in lower Transformer layers to maintain the base model's inherent language modeling capabilities. On the other hand, the upper layers employ bi-directional attention, allowing token-level decisions to incorporate both forward and backward contextual signals. This modification ensures the stable retention of linguistic knowledge and enhanced boundary detection  via bi-directional reasoning.

\noindent\textbf{Language Modeling as Token Critic:} Unlike conventional methods that predict token importance through scalar scores~\cite{jiang2023llmlinguacompressingpromptsaccelerated,pan2024llmlingua2datadistillationefficient} or restricted tagging schemes~\cite{li2023labelsupervisedllamafinetuning, dukic-snajder-2024-looking}, {\modelname} unifies semantic categorization and positional reasoning within a single framework.
As illustrated in Figure~\ref{fig:model_achitech}, we repurpose the base model's language modeling head to predict token-level semantic categories from the LLM vocabulary $\mathcal{V}$, while simultaneously employing a secondary head to generate sequence labeling scores for boundary detection. The resulting search space constitutes the Cartesian product of the vocabulary $\mathcal{V}$ the positional tag set. This combinatorial formulation expands the decision space by a considerable scale, enabling dynamic adaptation to diverse optimization criteria through prompt conditioning while preserving the base model's linguistic knowledge via parameter inheritance. With this language modeling as token critic setting, we maximize the log-probability of token's label during the supervised fine-tuning.

\noindent\textbf{Window-Parallel Inference: } We propose a window-parallel inference strategy for efficient context optimization, as depicted in Figure~\ref{fig:parrallel_compress}. Given a window size $w$, the long context $X_{l}$ is partitioned into $\lceil|X_{l}|/w\rceil$ non-overlapping windows. Each window is concatenated with the system prompt $P$ and query $q$ to form independent inputs for parallel processing via {\modelname}. Let $\rho$ denote the parallelism factor, the computational complexity of the LLM's pre-filling stage reduces to:
\begin{equation}\label{eq:complexity}
\text{O}(\frac{|X_l|}{\rho w}*w^2) + \text{O}(|X_s|^2) = \text{O}(\frac{w}{\rho}|X_l|) + \text{O}(|X_s|^2)
\end{equation}
In the above equation, $\text{O}(\frac{|X_l|}{\rho w}*w^2)$ represents the window-parallel inference overhead of {\modelname}, while $\text{O}(|X_s|^2)$  denotes the prefill computation cost for the generative LLM processing optimized context $X_s$. Given constant parameters $w$ and $\rho$, and $|X_s|\ll |X_l|$ in practice, this formulation achieves strictly lower complexity than the $\text{O}(|X_l|^2)$ baseline of direct prompting. The windowing strategy further enables theoretically infinite long context optimization - the length $|X_l|$ can scale with arbitrarily large while maintaining fixed memory overhead per parallel computation unit.

\subsection{Training Data Construction}\label{sec:data_construct}

The supervised fine-tuning process of {\modelname} is conducted based on the token critic task described in Section~\ref{sec:model_architecture}. To enhance contextual reasoning capabilities, we construct two specialized training datasets: (1) Multi-granularity context optimization data for fundamental context grounding skills, and (2) Query-aware context optimization data that specifically improves query-context interaction understanding. This dual training strategy enables simultaneous development of general contextual awareness and targeted query-response alignment capabilities.

\noindent\textbf{Multi-granularity context optimization: } In the multi-granularity context optimization training data, we focus on how {\modelname} can adaptively and controllably compress the long context into different granularities of optimized according to the system prompt:
\begin{itemize}[leftmargin=1em]
    \item \textbf{Keyword granularity}: We collected training data from open source datasets such as named entity recognition (NER)~\cite{conll03, conll04, levow-2006-third} and machine reading comprehension (MRC) with phrase-level answer~\cite{he-etal-2018-dureader,rajpurkar-etal-2018-know, kwiatkowski-etal-2019-natural}. In addition to these open datasets, we also crawled publicly accessible documents, including arXiv papers, financial reports, contracts, bidding announcements, and court judgments. We hired annotators to highlight and annotate keywords and entities within these documents and trained the model to extract these keywords.
    
    \item \textbf{Sentence granularity}: We modified the word granularity compression training data to sentence granularity compression, such as altering ``extract the keyword $k$ in the context'' to ``extract sentences mentioning the keyword $k$ in the context''. Additionally, we constructed a set of training data inspired by the ``Needle in a Haystack'' task~\cite{niah}, enabling the model to extract ``needle'' sentences from long documents. Furthermore, in sentence-level compression, we also aimed for {\modelname} to learn how to compress the long context into information-rich sentences. We assume that the summary of a context represents the richest information in that context to some extent. Therefore, we collected open-source summarization datasets~\cite{see-etal-2017-get} and our own annotated summarization datasets, leveraging the greedy search algorithm from SummaRuNNer~\cite{nallapati2016summarunnerrecurrentneuralnetwork} to construct sentence-level extractive summarization training data.
    
    \item \textbf{Paragraph granularity}: In paragraph-granularity compression training, we intended for {\modelname} to learn the ability to compress entire paragraphs. Using the document parsing toolkit DocMind~\footnote{\url{https://www.aliyun.com/product/ai/docmind}}, we segmented the crawled documents into paragraphs, tables, images, and other blocks. We employed annotators to label the meaning of randomly sampled paragraphs and tables, and trained the model to extract paragraphs or tables that correspond to these meanings from the context.
\end{itemize}

\noindent\textbf{Query-aware context optimization}: This training set focuses on enabling {\modelname} to perform context optimization aligned with user queries, thereby enhancing practical application performance. We leverage existing long-context QA datasets containing annotated supporting facts for reference answers~\cite{rajpurkar-etal-2018-know,hotpotqa,musique}, directly incorporating these into {\modelname}'s training to develop supporting fact extraction capabilities. To augment data diversity, we implement two complementary context compression synthesis approaches:

\begin{itemize}[leftmargin=1em]
\item \textbf{Forward synthesis} initiates with context segmentation into 256-token fragments, preserving semantic integrity through boundary sentence pruning. From these fragments, $N (1\leq N \leq 3)$ are randomly selected to generate query-answer pairs via self-instruction prompting~\cite{wang2023selfinstructaligninglanguagemodels}, with the LLM simultaneously producing supporting facts from the selected fragments. This process creates training instances for query-driven fact extraction from extended contexts.

\item\textbf{Backward synthesis} operates on pre-annotated query-answer pairs, applying identical context segmentation followed by map-reduce processing. The LLM evaluates each fragment's relevance to answering the target query, enabling construction of training data for context segment retrieval aligned with specific information needs.
\end{itemize}

Both synthesis pipelines incorporate answer consistency verification: generated supporting facts are fed back to the LLM for answer reproduction, with subsequent comparison against original annotations to filter low-quality outputs.

Through aformentioned methodologies, we curate a multi-domain, multilingual context compression corpus containing 126K samples (1.2B tokens) spanning various granularities and task types.

\section{Experimental Setup}\label{sec:detail}

\subsection{Implementation Details}

{\modelname} was initialized using the Qwen2-7b-Base architecture~\cite{yang2024qwen2technicalreport}, inheriting its parameters and vocabulary. The first 21 Transformer layers were retained as causal attention modules, while layers 22--28 were reconfigured as bi-directional location reasoning layers following the design in Section~\ref{sec:model_architecture}. A 3-epoch supervised fine-tuning regimen was implemented for the token critic task, employing the following configurations: window-parallel inference with 8192-token context windows, global batch size of 256, and constant learning rate of 1e-5. Training stability was ensured through Zero-3 partitioning with optimizer state offloading~\cite{rajbhandari2020zeromemoryoptimizationstraining}. To address input token optimization sparsity in long-context processing, we applied random gradient masking to 50\% of non-critical token positions during backpropagation. 


\subsection{Benchmarks}
Our evaluation protocol employs five long-context benchmarks with input lengths exceeding standard capabilities:

\noindent\textbf{Ruler-128K}~\cite{ruler}: Generates synthetic long-context data through controlled noise injection (essays, sentence repetitions, UUID strings) into conventional text. We assess performance on three representative subsets: Needle-in-a-Haystack (NIAH), Variable Tracking (VT), and Question Answering (QA). For cross-system comparability with closed-source approaches~\cite{MOBA}, we also select the multi-key level 2, multi-key level 3 and multi-value tasks from  Ruler-NIAH to derive the NIAH-sub benchmark.

\noindent\noindent\textbf{InfiniteBench}~\cite{infinitebench}: Features multilingual (English/Chinese) and multi-task evaluations across six categories: QA (EN/ZH), multiple choice (MC.EN) and specialized retrieval tasks (Passkey, Numeric, Key-Value). Context lengths span 122K--2M words, testing extreme-scale processing capabilities.

\noindent\textbf{Longbench V1}~\cite{bai2024longbench}: Comprehensive benchmark with 13K-word average context length. We evaluated the compared methods through established protocols~\cite{ye2025infiniteretrievalattentionenhanced,NSA} across three dimensions: single-document QA, multi-document QA, and summarization.

\noindent\textbf{Longbench V2}~\cite{bai2025longbenchv2deeperunderstanding}: Advanced successor featuring complex reasoning tasks with 8K-word to 2M-word contexts, requiring deep linguistic understanding and logical inference.

\noindent\textbf{Needle-in-a-Haystack}~\cite{niah}: Precision evaluation for target statement retrieval from irrelevant contexts. We test the performance of {\modelname} with the context lengths range from 32K to 1M tokens, and the inserted depths range from 0\% to 100\%.

\subsection{Compared Baselines}
Our comparative analysis includes state-of-the-art LLMs and specialized long-context management approaches across three categories:

\noindent\textbf{Proprietary LLMs}: Qwen2.5-Max~\cite{qwen2025qwen25technicalreport}, Qwen2.5-Turbo-1M~\cite{yang2025qwen251mtechnicalreport}, GPT-4o~\cite{openai2024gpt4ocard}, Claude-3.7-Sonnet~\cite{claude37}, Gemini-2.0-Pro~\cite{gemini2}, and Qwen-Long\footnote{\url{https://help.aliyun.com/zh/model-studio/long-context-qwen-long}}.

\noindent\textbf{Open-Source LLMs}: Qwen2.5-7b/32b/72b-instruct~\cite{qwen2025qwen25technicalreport}, DeepSeek-V3~\cite{deepseekai2025deepseekv3technicalreport}, LLaMA3.1-8b-instruct~\cite{grattafiori2024llama3herdmodels}, Qwen3-32b/235b-a22b~\cite{qwen3}.

\noindent\textbf{RAG}: The long context is partitioned into 600-token chunks with GTE embeddings~\cite{gteembedding} for top-k retrieval, accumulating results until reaching 16K tokens.

\noindent\textbf{Sparse Attention}: The open-source  implementation: Minference~\cite{jiang2024minference}. And the experimental results reported at the original papers of the established methods: InfiniteRetrieval~\cite{ye2025infiniteretrievalattentionenhanced}, MOBA~\cite{MOBA}, and NSA~\cite{NSA}.

\section{Experimental Results}\label{sec:expriment}

%




%

\begin{table*}[t]
    \caption{Evaluation results on Ruler-128K and InfiniteBench}
    \label{tab:main_res_ruler_and_infbench}
    \footnotesize
    \centering
    \resizebox{\linewidth}{!}{
    \begin{tabular}{l|ccccc|ccccccc}
    \toprule
    \multirow{2}{*}{\textbf{Model}} & \multicolumn{5}{c|}{\textbf{Ruler-128k}} & \multicolumn{7}{c}{\textbf{InfiniteBench}} \\
    ~ & \textbf{NIAH-sub}& \textbf{NIAH} & \textbf{VT} & \textbf{QA}& \textbf{Avg} & \textbf{QA.EN} & \textbf{QA.ZH}  
    & \textbf{MC.EN} & \textbf{RT.Passkey} & \textbf{RT.NUM} & \textbf{RT.KV}& \textbf{Avg}  \\
    \midrule
    Qwen2.5-72b-instruct~\cite{qwen2025qwen25technicalreport}& 47.92&75.00&91.60&66.00&77.53 &27.36&31.29&76.42&79.66&79.66&32.00&54.40 \\
    Qwen2.5-max~\cite{qwen2025qwen25technicalreport}&10.00 &19.84&23.60&38.00&27.15  &18.11&18.87&69.87&22.03&22.33&16.40&27.93 \\
    Qwen2.5-turbo-1M~\cite{yang2025qwen251mtechnicalreport}& 41.67&75.00&90.00&65.31& 76.77&17.46&25.86&65.94&\textbf{100.00}&99.83&38.00&57.85\\
    Qwen3-32b&48.26 &60.54 &70.08 &40.37 &57.00 &15.83&26.86&75.55&71.19&71.19&32.00&48.77\\
    Qwen3-235b-a22b&48.60 &61.57 &70.04 &39.55 &57.05 &9.62&17.73&77.29&71.19& 71.19&33.00&46.67\\
    gpt-4o~\cite{openai2024gpt4ocard}&47.92 & 71.25&77.60&\textbf{76.00}&74.95&31.72&25.24&86.46&79.66&79.66&58.00&60.12\\
    claude-3.7-sonnet~\cite{claude37}&49.17 &70.94&76.00&70.00& 72.31&35.95&17.02&74.67&79.62&80.00&\textbf{59.20}&57.74 \\
    deepseek-v3~\cite{deepseekai2025deepseekv3technicalreport}&16.25 &30.47&34.40&30.00& 31.26&20.69&26.76&74.67&41.19&41.86&21.60&37.18\\
    gemini-2.0-pro~\cite{gemini2}&58.47 & 71.63&75.48&72.20 & 73.10&\textbf{38.06}&\textbf{32.03}&\textbf{90.39}&79.66&79.66&57.80&\textbf{62.93} \\
    Qwen-Long& \textbf{86.61}&\textbf{94.92}&\textbf{99.84}&68.70&\textbf{87.82} & 17.46& 25.80& 71.56 & \textbf{100.00} & \textbf{100.00}& 36.20& 58.50\\
    \midrule
    LLaMA3.1-8b-instruct~\cite{grattafiori2024llama3herdmodels}&47.22 &65.69&28.63&59.80&51.37&26.74&35.18&26.20&94.58&91.86&64.00&56.43 \\
    \ \ \  + RAG~\cite{gteembedding}&78.40& 91.20& 93.32& 71.80& 85.44& 23.22&35.78&34.70&96.70& 93.60&68.18& 58.70 \\
    \ \ \ + MOBA~\cite{MOBA}&75.58&-&-&-&-&-&-&-&-&-&-&- \\
    \ \ \ + MInference~\cite{jiang2024minference}&47.80 &73.66&29.08&46.10& 49.16&\textbf{27.60}&30.28&44.54&87.80&84.24&22.60&49.51 \\ 
    \rowcolor{gray!10}\ \ \ + {\modelname}&\textbf{99.65} &\textbf{99.75}&\textbf{100.00}&\textbf{73.53}& \textbf{91.09}&26.20&\textbf{35.82}&\textbf{64.63}&\textbf{95.93}&\textbf{96.78}&\textbf{99.00}&\textbf{69.73}\\ 
    \midrule
    Qwen2.5-7b-instruct~\cite{qwen2025qwen25technicalreport}&21.33&59.98&24.48&45.90 &43.45 & 16.69 & 21.34 & 59.82 & 94.91 & 94.91 & 14.60&50.38\\
    \ \ \ + RAG~\cite{gteembedding}&  82.65&93.30& 98.92&65.70&85.97 &15.56&20.26&79.04&100.00&99.66&73.60&64.69 \\
    \ \ \ + MInference~\cite{jiang2024minference}&37.87 &69.34&22.92&45.70& 45.99&16.80&21.12&62.54&94.92&94.92&13.20&50.38\\
    \rowcolor{gray!10}\ \ \ + {\modelname}& \textbf{99.87}& \textbf{99.85} & \textbf{98.68} & \textbf{72.20}&\textbf{90.24} & \textbf{26.43} & \textbf{28.82} & \textbf{79.91} & \textbf{100.00} & \textbf{100.00} & \textbf{98.80}&\textbf{72.33}\\
    \midrule
    Qwen2.5-32b-instruct~\cite{qwen2025qwen25technicalreport}&50.70 & 75.76 & 93.56 & 50.61&73.31  & 20.88 & 18.17 & 79.04 & 91.53 & 85.08 & 35.20&54.98 \\
    \ \ \ + RAG~\cite{gteembedding}&86.41  &93.38&97.92&65.30& 85.53&14.97&20.44&76.86&99.66&100.00&73.00&64.16 \\
    \ \ \ + MInference~\cite{jiang2024minference}&50.62 &76.14&95.84&51.10&74.36&21.65&18.0&80.79&94.92&94.92&36.4&57.78 \\
    \rowcolor{gray!10} \ \ \ + {\modelname} &\textbf{99.93} & \textbf{99.95} & \textbf{99.86} & \textbf{78.20} &\textbf{92.67}& \textbf{26.63}&\textbf{26.62}&\textbf{90.39}&\textbf{100.00}&\textbf{100.00}&\textbf{99.20} &\textbf{73.81} \\
    \bottomrule
    \end{tabular}
    }
    \vspace{-0.5cm}
\end{table*}

\begin{table*}[t]
    \caption{Evaluation results on Longbench V1}
    \label{tab:main_res_longbench_v1}
    \footnotesize
    \centering
    \resizebox{\linewidth}{!}{
    \begin{tabular}{l|cccc|cccc|c|cccc|c}
    \toprule
    \multirow{2}{*}{\textbf{Model}} & \multicolumn{4}{c|}{\textbf{SingleDoc QA}} & \multicolumn{4}{c|}{\textbf{MultiDoc QA}}&\multirow{2}{*}{\textbf{Avg\_qa}}&\multicolumn{4}{c|}{\textbf{Summary}} &\multirow{2}{*}{\textbf{Avg}}\\
    ~ & \textbf{MF (en)} & \textbf{\textbf{MF (zh)}} & \textbf{NQ}& \textbf{Qasp} & \textbf{HPQA} & \textbf{2Wiki.}  
    & \textbf{Musi.} & \textbf{Du.}& & \textbf{GR} & \textbf{QM}& \textbf{MN}&\textbf{VC}&  \\
    \midrule
    Qwen2.5-72b-instruct~\cite{qwen2025qwen25technicalreport}& 54.26&66.20&33.78&47.56&65.88&64.46&42.00&19.29&45.89&19.56&18.69&13.52&16.30&38.46\\
    Qwen2.5-max~\cite{qwen2025qwen25technicalreport}& 51.40&64.75&27.80&43.47&65.77&62.51&40.16&20.09& 46.99&16.90&18.64&13.57&16.52&36.80\\
    Qwen2.5-turbo-1M~\cite{yang2025qwen251mtechnicalreport}&54.41&60.66&24.59&42.73&63.03&52.72&39.21&\textbf{23.88} & 45.15&17.33&18.39&14.32&\textbf{19.12}&35.87\\
    Qwen3-32b&46.71&62.90&27.79&40.73&66.56&74.00&52.57&18.48 &48.72 &27.73&19.93&21.69&13.99&39.42\\
    Qwen3-235b-a22b& 45.60&61.45&30.16& 42.06&68.40&75.42&56.20&17.59&49.61&27.35&20.00&21.45&13.50&39.93\\
    gpt-4o~\cite{openai2024gpt4ocard}&53.99&62.56&35.13&46.97&68.79&68.51&46.20&17.71&49.98 &15.82&17.11&14.40&14.95&38.51 \\
    claude-3.7-sonnet~\cite{claude37}&53.45&63.82&\textbf{37.57}&48.80&66.15&67.22&48.37&19.53&50.16 &18.08&16.73&15.42&16.61&39.61 \\
    deepseek-v3~\cite{deepseekai2025deepseekv3technicalreport}&\textbf{56.69}&\textbf{66.72}&34.80&47.59&69.53&69.65&57.62&21.23&52.98 &17.85&17.85&15.06&17.81&41.03\\
    gemini-2.0-pro~\cite{gemini2}& 55.77&65.47&33.10&\textbf{51.39}&\textbf{70.69}&\textbf{80.43}&\textbf{68.18}&12.10& \textbf{54.64}&16.41&17.41&13.35&10.64&\textbf{41.25}\\
    Qwen-Long& 53.62&64.49&30.92&43.98&64.80&62.53&41.06&21.83& 47.90&\textbf{33.04}&\textbf{24.06}&\textbf{22.44}&13.71&39.71\\
    \midrule
    LLaMA3.1-8b-instruct~\cite{grattafiori2024llama3herdmodels}&49.39&60.64&26.81&37.79&53.07&40.42&29.79&21.06& 39.87&19.09&17.99&14.80&9.63&31.71 \\
    \ \ \ + RAG~\cite{gteembedding} &32.09&25.91&12.83&18.36&30.91&21.35&10.64&15.21&20.91 &12.09&15.50&13.77&6.94 &17.97\\
    \ \ \ + MInference~\cite{jiang2024minference}&\textbf{55.21}&60.69&28.61&\textbf{47.57}&53.60&40.43&29.08&\textbf{27.76}& 42.87&19.88&18.19&15.40&\textbf{17.37}&34.48 \\
    \ \ \ + InfiniteRetrieval~\cite{ye2025infiniteretrievalattentionenhanced}&44.72&- &18.88&36.45& 50.10&29.98&27.26&-& -&\textbf{21.94}&\textbf{20.17}&\textbf{24.14}&-&- \\
    \rowcolor{gray!10}\ \ \ + {\modelname}&54.37&\textbf{60.87}&\textbf{28.68}&42.92&\textbf{59.03}&\textbf{52.46}&\textbf{34.58}&23.93&\textbf{44.60} &20.02&18.22&15.06&16.84&\textbf{35.58} \\
    \midrule
    Qwen2.5-7b-instruct~\cite{qwen2025qwen25technicalreport}&52.35&62.94&25.40&45.40&56.92&45.02&31.16&\textbf{28.57}& 43.47&\textbf{32.64}&\textbf{23.35}&\textbf{23.20}&13.69&\textbf{36.72} \\
    \ \ \ + RAG~\cite{gteembedding} & 33.93&26.97&12.15&21.97&33.15&23.72&11.14&20.64& 22.96&20.48&20.15&21.41&9.81&21.29\\
    \ \ \ + MInference~\cite{jiang2024minference}&51.86&\textbf{62.95}&\textbf{28.25}&\textbf{45.87}&56.29&43.03&31.16&23.15&42.82&18.57&18.64&14.50&17.89&34.35 \\
    \ \ \ + InfiniteRetrieval~\cite{ye2025infiniteretrievalattentionenhanced}&50.92& -& 25.48& 42.12& 57.52&50.26&30.62 &-&-& 19.26&20.47& 20.60&-&- \\
    \rowcolor{gray!10}\ \ \ + {\modelname}& \textbf{53.42}&62.48&24.97&44.16& \textbf{63.95}&\textbf{56.16}&\textbf{41.01}&19.04&\textbf{45.65} &17.37&17.63&15.30&\textbf{18.02}&36.12\\
    \midrule
    Qwen2.5-32b-instruct~\cite{qwen2025qwen25technicalreport} & 52.68&\textbf{66.22}&\textbf{31.91}&\textbf{47.99}&64.81&59.81&40.60&\textbf{22.00}& 48.25&17.09&17.84&14.88&18.80&37.89 \\
    \ \ \ + RAG~\cite{gteembedding} & 34.17&32.17&13.21&22.79&35.86&25.30&13.19&21.13&24.73&\textbf{27.83}&\textbf{20.12}&\textbf{21.29}&10.15&23.10\\
    \ \ \ + MInference~\cite{jiang2024minference}&50.98&66.14&27.90&46.46&62.37&59.22&39.13&24.90&47.17&18.46&17.62&14.32&17.36&37.07\\
    \rowcolor{gray!10} \ \ \ + {\modelname} &\textbf{54.59}&65.82& 30.56&45.54&\textbf{67.44}&\textbf{66.28}&\textbf{47.89}&19.81& \textbf{49.74}&16.37&18.01&14.62&\textbf{19.23}&\textbf{38.85} \\
    \bottomrule
    \end{tabular}
    }
    \vspace{-0.2cm}
\end{table*}

\begin{table*}[t]
    \caption{Evaluation results on Longbench V2}
    \label{tab:main_res_longbench_v2}
    \footnotesize
    \centering
    \resizebox{0.85\linewidth}{!}{
    \begin{tabular}{l|c|cc|ccc}
    \toprule
    \multirow{2}{*}{\textbf{Model}}
    & \multirow{2}{*}{\textbf{Overall}}&\multicolumn{2}{c|}{\textbf{Difficulty}}&\multicolumn{3}{c}{\textbf{Length (< 32K; 32K-128K; > 128K)}}\\
    &&\textbf{Easy}&\textbf{Hard}&\textbf{Short}&\textbf{Medium}&\textbf{Long }\\
    \midrule
    Qwen2.5-72b-instruct~\cite{qwen2025qwen25technicalreport} & 42.1 & 42.7 & 41.8 & 45.6 & 38.1 & 44.4 \\
    Qwen2.5-max~\cite{qwen2025qwen25technicalreport} & 46.5 & 51.6 & 43.4 & 56.7 & 38.6 & 45.4 \\
    Qwen2.5-turbo-1M~\cite{yang2025qwen251mtechnicalreport} & 40.8 & 45.3 & 37.9 & 46.1 & 37.7 & 38.0\\
    Qwen3-32b~\cite{qwen3}& 48.7& 57.3&43.4&57.2&42.3&47.2\\
    Qwen3-235b-a22b~\cite{qwen3}& 51.9&58.1&48.1&60.3&46.7&48.1\\
    gpt-4o~\cite{openai2024gpt4ocard} & 46.0 & 50.8	& 43.0 & 47.5 & 47.9 & 39.8 \\
    claude-3.7-sonnet~\cite{claude37} & 50.9 & 60.4 & 45.0 & 55.0 & 47.9 & 50.0 \\
    deepseek-v3~\cite{deepseekai2025deepseekv3technicalreport} & 45.3 & 51.0 & 41.8 & 51.7 & 40.0 & 45.4 \\
    gemini-2.0-pro~\cite{gemini2} & \textbf{60.6} & \textbf{71.6} & \textbf{53.7} & \textbf{69.1} & \textbf{54.0} & \textbf{59.4} \\
    Qwen-Long&43.0& 49.2& 39.2&48.3&37.7&44.9\\
    \midrule
    Qwen2.5-7b-instruct~\cite{qwen2025qwen25technicalreport} & 27.4 & 30.2 & 25.7 & 35.0 & 24.2 & 21.3 \\
    \ \ \ + RAG~\cite{gteembedding} & 31.4 & \textbf{35.9} & 28.6 & 33.3 & 28.4 & \textbf{34.3} \\
    \ \ \ + MInference~\cite{jiang2024minference} & 27.8 & 27.6 & 28.0 & 31.7 & 27.0 & 23.1 \\
    \rowcolor{gray!10}\ \ \ + {\modelname} & \textbf{33.1} & 30.9 & \textbf{34.4} & \textbf{36.7} & \textbf{29.8} & 33.6 \\
    \midrule
    Qwen2.5-32b-instruct~\cite{qwen2025qwen25technicalreport} & 41.7 & \textbf{47.9} & 37.9 & \textbf{46.1} & \textbf{38.1} & 41.7  \\
    \ \ \ + RAG~\cite{gteembedding} & 32.2 & 36.5 & 29.6 & 35.6 & 27.0 & 37.0 \\
    \ \ \ + MInference~\cite{jiang2024minference} & 35.8 & 38.0 & 34.4 & 32.8 & \textbf{38.1} & 36.1 \\
    \rowcolor{gray!10} \ \ \ + {\modelname} & \textbf{42.0} & 45.5 & \textbf{39.9} & 42.2 & \textbf{38.1} & \textbf{49.5}  \\
    \bottomrule
    \end{tabular}
    }
\vspace{-0.5cm}
\end{table*}

\subsection{Overall Results}\label{sec:main_res}
The evaluation results across Tables~\ref{tab:main_res_ruler_and_infbench}, \ref{tab:main_res_longbench_v1}, and \ref{tab:main_res_longbench_v2} demonstrate {\modelname}'s effectiveness against comparative baselines, revealing several key findings:


\noindent\textbf{Performance Enhancement Through {\modelname}}: Model cascading with {\modelname} yields consistent improvements across all evaluated LLMs. LLaMA3.1-8b-Instruct, Qwen2.5-7b-Instruct, and Qwen2.5-32b-Instruct achieve respective performance gains of 39.72, 55.79, and 19.26 on Ruler-128K, with comparable improvements of 13.30, 21.95, and 18.83 on InfiniteBench. LongBench evaluations show sustained enhancements, averaging +2.8 on V1 and +3.0 on V2. 
 Significantly, {\modelname}-augmented open-source models surpass proprietary LLMs on Ruler-128K and InfiniteBench, establishing new state-of-the-art performances. This demonstrates that \emph{resource-constrained open-source LLMs can achieve parity with commercial counterparts in long-context tasks when integrated with {\modelname}.}


\noindent\textbf{Superior Performance with Extended Context Lengths}: Experimental results demonstrate {\modelname}'s effectiveness correlates positively with input context length across evaluated tasks. It shows pronounced advantages when processing contexts exceeding standard LLM capacity limits, achieving average performance gains of 38.20 on Ruler-128K, 18.02 on InfiniteBench, and 10.5 on LongBench V2-Long. Conversely, minor improvements occur on LongBench V1's shorter contexts (2K-18K tokens). This dichotomy reveals two critical insights: (1) Current LLMs exhibit sufficient competence for conventional-length tasks, and (2) {\modelname} provides essential performance augmentation specifically for extreme-length scenarios beyond standard model capacities. These findings strategically position {\modelname} as a solution for ultra-long context applications.

\noindent\textbf{Multi-Granularity Context Optimization}: As detailed in Section~\ref{sec:method}, {\modelname} enables dynamic context optimization through system-prompt-controlled granularity adaptation. Table~\ref{tab:res_gran_and_prompt} presents the implemented prompt-granularity configurations for Ruler-128K and InfiniteBench tasks, demonstrating consistent performance gains across all granularity levels. This multi-scale optimization capability permits customized prompt engineering while maintaining robust performance, which is particularly valuable for applications requiring flexible context optimization strategies.

\begin{table*}[t]
    \caption{The system prompt and performances of {\modelname} in different Tasks, with Qwen2.5-7b-instruct as the generative LLM. The ``Avg. Len.'' represents the average token numbers of the optimized context.}
    \label{tab:res_gran_and_prompt}
    \footnotesize
    \centering
    \resizebox{\linewidth}{!}{%
    \begin{tabular}{l|c|c|c|>{\raggedright\arraybackslash}p{0.45\linewidth}}
    \toprule
     \textbf{Task} & \textbf{Granularity} & \textbf{Avg. Len.}&\textbf{Improvement} & \textbf{Prompt Example} \\
    \midrule
    InfiniteBench-RT.KV & Words (UUID) & 69.32 & +84.20& \multirow{2}{\hsize}{\raggedright\texttt{Extract the \{key\}:\{value\} pair for the key in user's question}}\\
    & & & &\\
    \midrule
    Ruler-128K-NIAH & \multirow{4}{*}{Short Sentences} & 52.13 & +39.87 & \multirow{4}{\hsize}{\raggedright\texttt{Extract the `needles' in the format of `One of the special magic \{type\_needle\_v\} for \{key\} is: \{value\}.' from the document}} \\ 
    Ruler-128K-VT & & 53.30 & +74.20 & \\ 
    InfiniteBench-RT.Passkey &&24.00 &+5.19 & \\ 
    InfiniteBench-RT.NUM & &36.95 &+5.19 & \\ 
    \midrule
    & & & &\multirow{3}{\hsize}{\raggedright\texttt{Extract some sentences from the documents as the supporting facts for answering the user's question.}}\\
    Ruler-128K-QA & Long Sentences& 1173.17& +13.73& \\
    & & & &\\
    \midrule
    InfiniteBench-EN.QA&\multirow{3}{*}{Long Paragraphs} & 26287.13 & +9.74&\multirow{3}{\hsize}{\raggedright\texttt{Retrieve chunks or paragraphs from the document that are related to the user's query.}}\\
    InfiniteBench-ZH.QA && 63737.20&+7.48 & \\
    InfiniteBench-EN.MC& & 30466.04& +20.09&\\
    \bottomrule
    
    \end{tabular}%
    }
    \vspace{-0.2cm}
\end{table*}

\begin{table*}[t]
    \caption{Performance comparison between {\modelname} and NSA}
    \label{tab:comparison_nsa}
    \footnotesize
    \centering
    \resizebox{\linewidth}{!}{
    \begin{tabular}{l|ccc|cccc|l}
    \toprule
    \multirow{2}{*}{\textbf{Model}} & \multicolumn{3}{c|}{\textbf{SingleDoc QA}} & \multicolumn{4}{c|}{\textbf{MultiDoc QA}}&\multirow{2}{*}{\textbf{Avg ($\Delta$)}}  \\
    ~ & \textbf{MF (en)} & \textbf{\textbf{MF (zh)}} & \textbf{Qasp} & \textbf{HPQA} & \textbf{2Wiki.}  
    &\textbf{Musi.}& \textbf{Du.}&\\
    \midrule
    DeepSeekMOE~\cite{dai-etal-2024-deepseekmoe}&51.20& 62.30& 40.90&35.00&30.50&32.40&29.40&40.24\\
    \ \ \ + NSA~\cite{NSA}& 50.30&62.40&43.20&43.70&35.60&30.70&34.10&42.86 (+2.62)\\
    \midrule
    LLaMA3.1-8b-instruct~\cite{grattafiori2024llama3herdmodels}&49.39&60.64&37.79&53.07&40.42&29.79&21.06&41.74  \\
    \rowcolor{gray!10}\ \ \ + {\modelname}&54.37&60.87&42.92&59.03&52.46&34.58&23.93&46.88 (+\textbf{5.14}) \\
    \bottomrule
    \end{tabular}
    }
    \vspace{-0.5cm}
\end{table*}


\noindent\textbf{Comparative Analysis with RAG}: The evaluation reveals distinct performance characteristics between {\modelname} and RAG approaches. RAG demonstrates effectiveness in elementary retrieval tasks requiring single-fact extraction from noisy contexts (e.g., Ruler-NIAH, RT.Passkey, RT.NUM), achieving 82.4\% average accuracy across these benchmarks. However, its performance degrades markedly in complex scenarios: (1) multi-hop reasoning tasks (QA.EN, QA.ZH, MultiDoc QA) and contexts with high query-context similarity (RT.KV). In contrast {\modelname} effectively mitigates these limitations through dynamic context optimization, achieving 18.52\% and 25.20\% relative improvements respectively in these challenging scenarios. 

\noindent\textbf{Comparative Analysis with Sparse Attention}: Our analysis contrasts {\modelname} against two sparse attention paradigms: training-free approaches (Minference~\cite{jiang2024minference}, InfiniteRetrieval~\cite{ye2025infiniteretrievalattentionenhanced}) and trainable implementations (MOBA~\cite{MOBA}, NSA~\cite{NSA}). 
The training-free methods demonstrate limited improvements on Ruler-128K and Infinitebench, with Minference underperforming direct prompting on other benchmarks. The trainable MOBA achieves +28.36 on NIAH-sub versus direct prompting, though substantially lower than {\modelname}'s +52.43 gain. 
For NSA which is trained with DeepseekMOE~\cite{dai-etal-2024-deepseekmoe}), relative improvements with comparable base models in Table~\ref{tab:comparison_nsa} reveal {\modelname}'s superior scalability (+5.14 vs NSA's +2.62). This comparative analysis demonstrates that while parametric sparse attention requires model-specific training for marginal gains, while {\modelname} delivers performance advantages across heterogeneous architectures without specialized optimization.

 \begin{figure*}[t]
    \centering
    \includegraphics[width=1.0\linewidth]{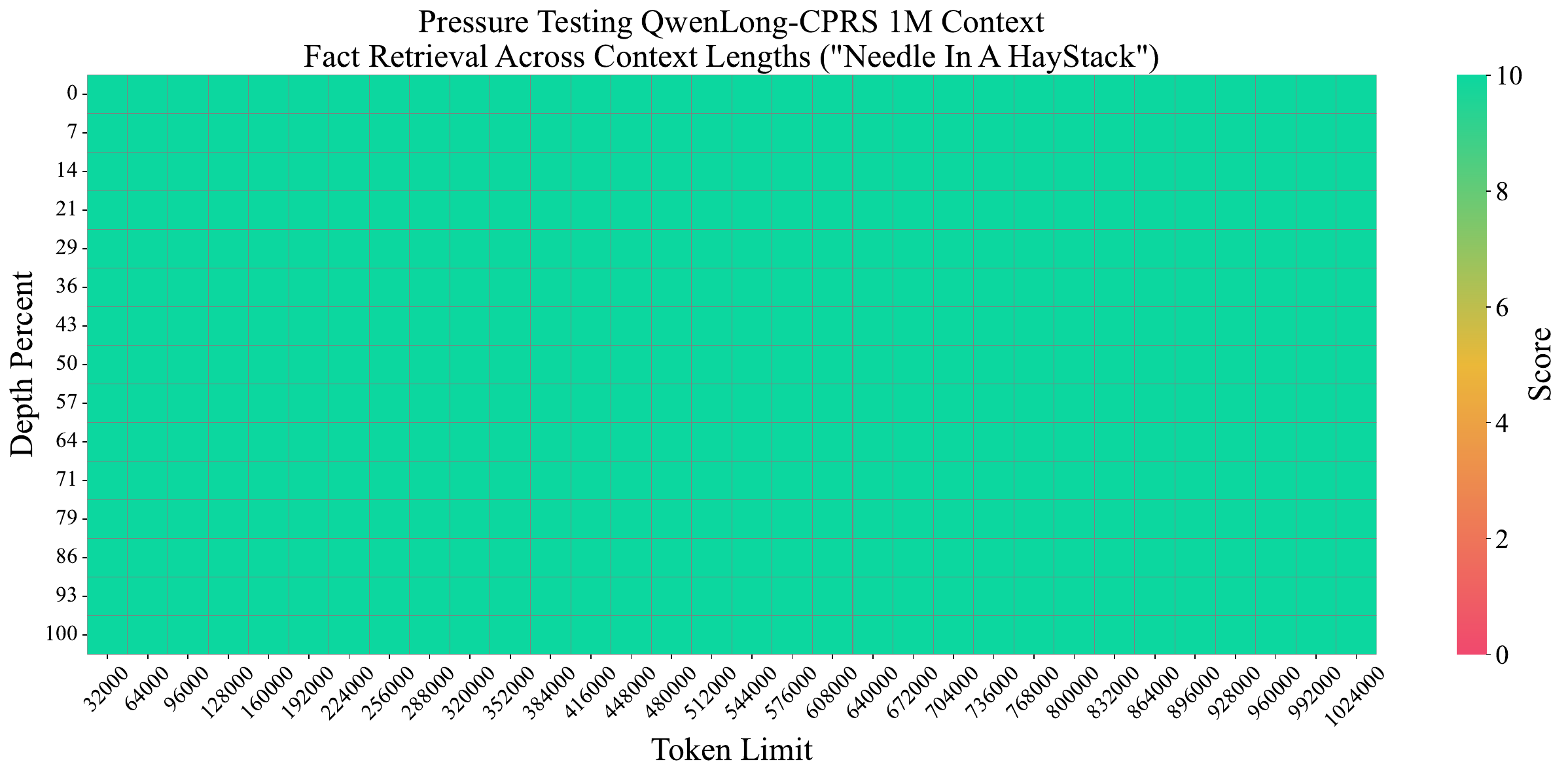}
    \vspace{-0.5cm}
    \caption{Performance of {\modelname} in NIAH test with input length is upto 1M.} 
    \label{fig:niah}
\end{figure*}

\noindent\textbf{Depth-Robust Needle Retrieval}: Figure~\ref{fig:niah} demonstrates {\modelname}-enhanced Qwen2.5-7b-Instruct's performance in the Needle-in-a-Haystack paradigm across full-spectrum depth variations (0\% to 100\%) and context lengths (32K to 1M tokens). The system achieves perfect accuracy scores under all test configurations, matching the claimed capabilities of contemporary LLMs and agent systems advertising over 1M token capacities~\cite{MOBA, qwen2025qwen25technicalreport, yang2025qwen251mtechnicalreport, qwen-agent, minimax2025minimax01scalingfoundationmodels}.

 \begin{figure*}[t]
    \centering
    \includegraphics[width=1.0\linewidth]{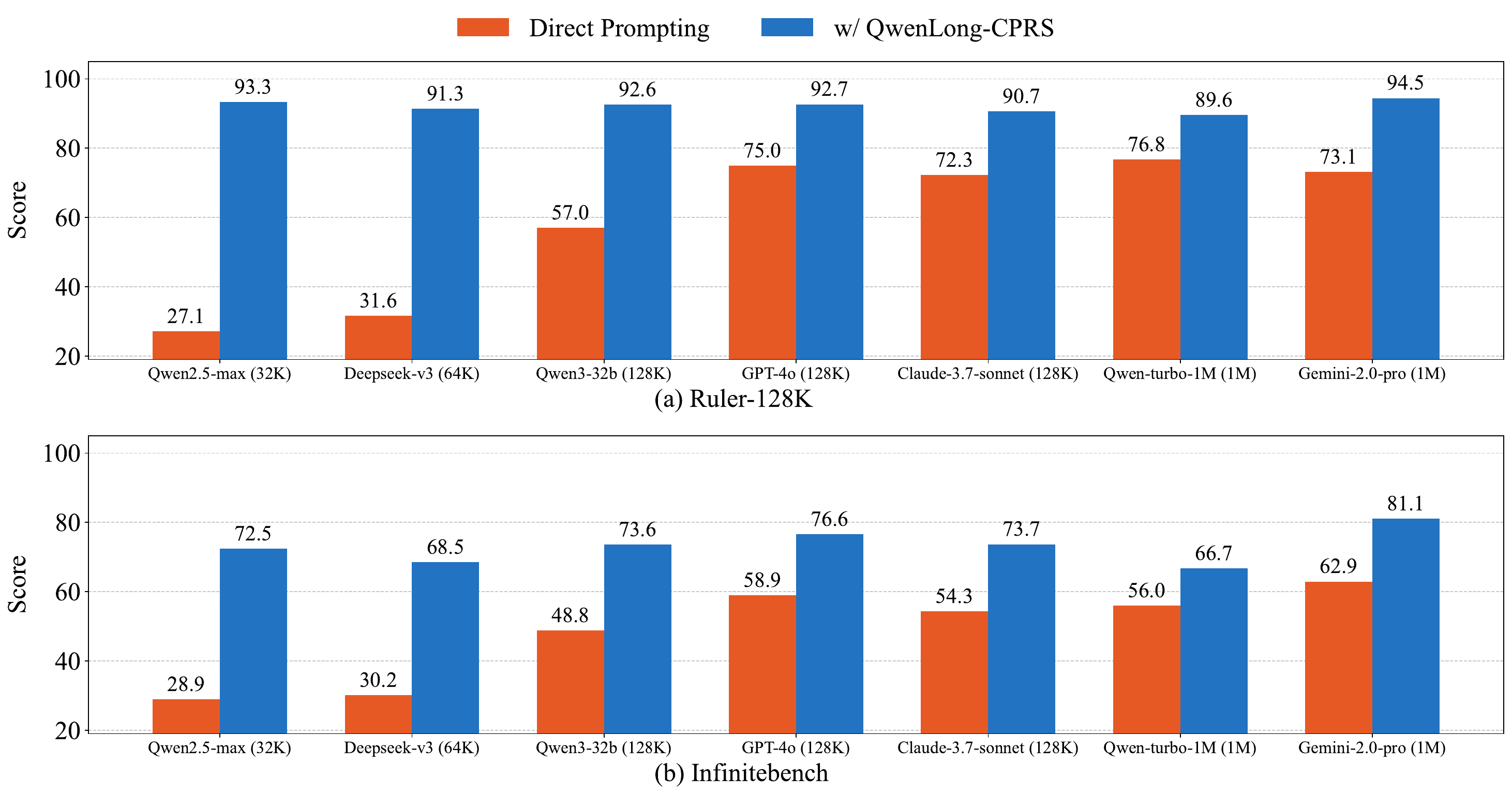}
    \vspace{-0.5cm}
    \caption{Comparative performance analysis of LLMs with and without {\modelname} integration. Numerical values in parentheses indicate each model's maximum input capacity.} 
    \label{fig:llm_and_compress}
\end{figure*}

\subsection{{\modelname} with Stronger LLMs}

To investigate {\modelname}'s capacity to push performance boundaries in long-context scenarios, we deploy it across high-parameter LLMs with superior general capabilities. Figure~\ref{fig:llm_and_compress} reveals consistent performance enhancements: all {\modelname}-augmented models exceed prior direct prompting baselines with the best performance (77.5 on Ruler-128K and 62.9 on InfiniteBench).

Crucially, {\modelname} effectively compensates for varying input constraints\footnote{For Qwen2.5-max and Deepseek-v3, we utilize the version provided by the Aliyun Bailian platform, which supports 32K and  64K input contexts, respectively.}, delivering average gains of 54.9, 49.0, 21.7 and 15.7 for the LLMs with 32K, 64K, 128K and 1M context windows, respectively. This demonstrates {\modelname}'s ability to elevate length-limited LLMs to parity with specialized long-context LLMs through optimized context, enabling resource-efficient deployment of short-input LLMs in extended-sequence applications.

\begin{figure*}[t]
\centering
\includegraphics[width=1.0\linewidth]{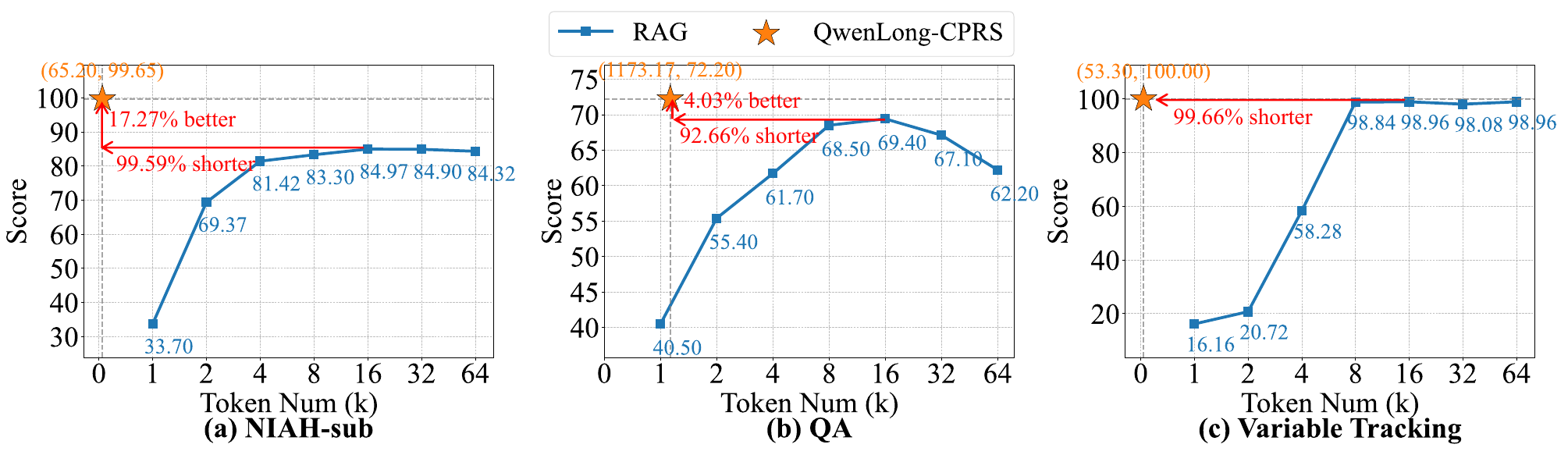}
\vspace{-0.5cm}
\caption{Performance comparison between RAG and {\modelname} across varying retrieved token quantities}
\label{fig:rag_vs_compress}
\end{figure*}
\subsection{Context Optimization Efficiency}\label{sec:retrieve_and_compress}
This section analyzes token efficiency by examining RAG's performance variation with increasing retrieved tokens from 1K to 64K across three Ruler-128K subsets. As shown in Figure~\ref{fig:rag_vs_compress}, RAG exhibits suboptimal performance when the retrieved tokens are less than 8K across all tasks due to critical information loss in coarse-grained retrieval. RAG's performance peaks at 16K tokens, and then keeps fluctuation on NIAH-sub and Variable Tracking, and even go worse on QA when the number of retrieved tokens increase. This pattern emerges from RAG's incomplete critical data capture at low token quantities and noise accumulation at higher scales.

{\modelname} demonstrates superior performance through context optimization, achieving 99.59\%, 92.66\%, and 99.66\% less tokens than RAG's peak-requirement volumes the three tasks. Furthermore, it surpasses RAG's maximum scores by 17.27\% on NIAH-sub and 4.03\% on QA, establishing both quantitative and qualitative advantages in context optimization efficiency.

 \begin{figure*}[t]
    \centering
    \includegraphics[width=0.9\linewidth]{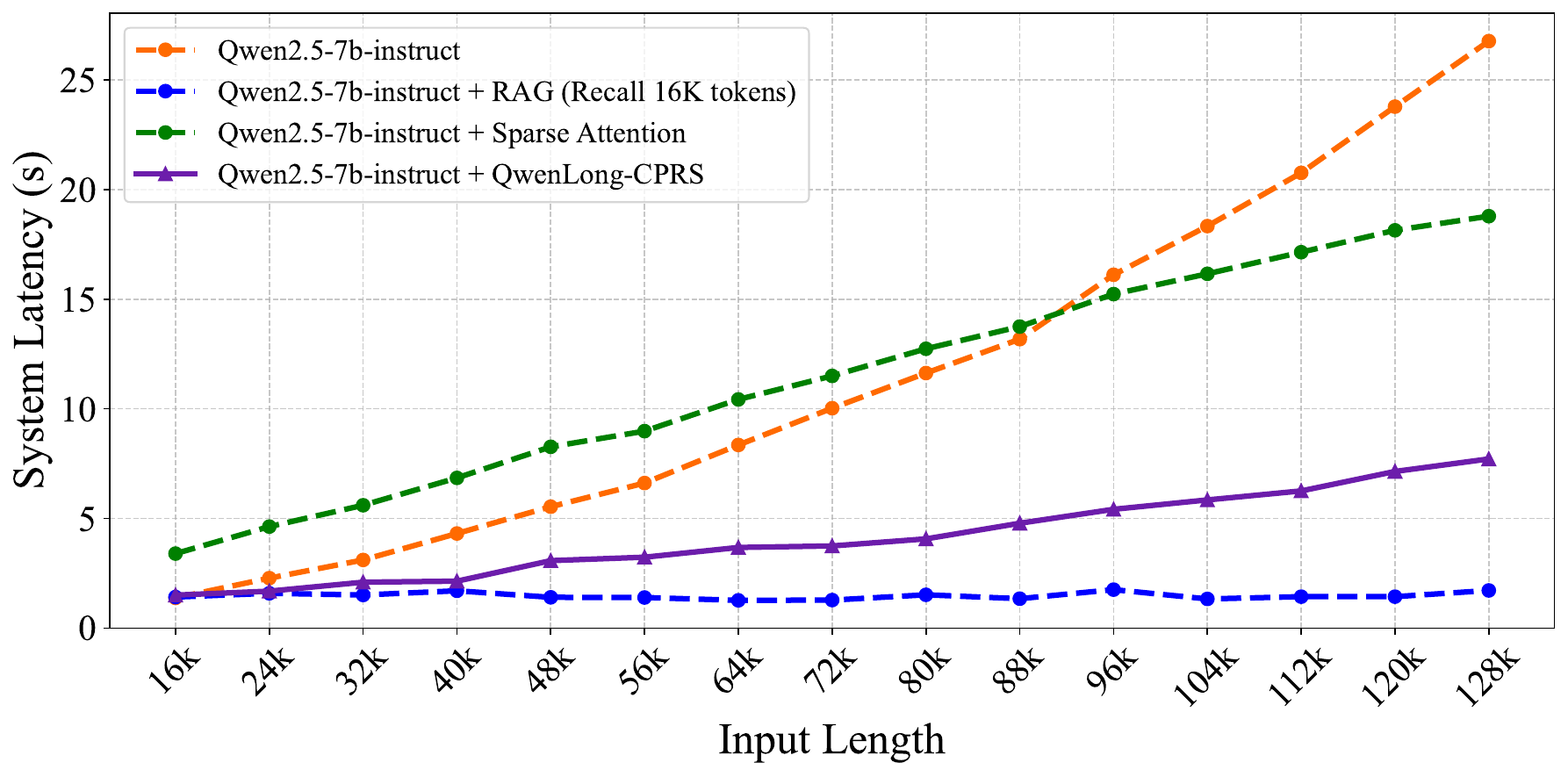}
    \vspace{-0.2cm}
    \caption{System latency of different context management methods with various input length.} 
    \label{fig:time_test}
\end{figure*}

\subsection{Latency Analysis}
We evaluate {\modelname}'s impact on LLM prefill latency through four system configurations: (1) Direct prompting with Qwen2.5-7b-instruct, (2) RAG-enhanced implementation (16K retrieved tokens), (3) Minference sparse attention integration, and (4) {\modelname}-cascaded architecture illustrated in Figure~\ref{fig:parrallel_compress}. The {\modelname} configuration employs window size $w=8192$ and parallelism factor $\rho=5$. All systems utilize Qwen2.5-7b-instruct deployed 1 NVIDIA A100 GPU via \texttt{vLLM} with paged attention~\cite{vllm} for memory optimization. Figure~\ref{fig:time_test} presents Time-to-First-Token (TTFT) measurements of the above four systems under increasing context lengths, revealing three critical patterns:
\begin{itemize}[leftmargin=1em]
    \item The baseline direct prompting method exhibits quadratic latency growth, reaching 26.76s at 128K tokens. {\modelname} demonstrates superior linear scaling with 3.47$\times$ acceleration over baseline at 128K, despite current implementation limitations that leave the optimization of computation kernel unexplored.
    \item Minference's sparse attention shows paradoxical behavior: below 96K tokens, dynamic sparse pattern computation overhead causes higher latency than direct prompting (10.42s vs 8.35s at 64K). At 128K tokens, it achieves 1.42$\times$ acceleration, which is substantially lower than {\modelname}'s 3.47$\times$ improvement.
    \item RAG demonstrates constant-time latency characteristics, though this comes at accuracy costs shown in Section~\ref{sec:main_res} and Section~\ref{sec:retrieve_and_compress}. Our future work will focus on bridging {\modelname}'s current linear complexity toward constant-time performance while preserving its accuracy advantages.
\end{itemize}

\begin{table*}[t]
    \caption{Performance Comparison of {\modelname}-Augmented LLMs with Original vs. Customized Prompts}
    \label{tab:res_different_prompt}
    \footnotesize
    \centering
    \resizebox{\linewidth}{!}{
    \begin{tabular}{l|cccc|cccc|c|cccc|c}
    \toprule
    \multirow{2}{*}{\textbf{Model}} & \multicolumn{4}{c|}{\textbf{SingleDoc QA}} & \multicolumn{4}{c|}{\textbf{MultiDoc QA}}&\multirow{2}{*}{\textbf{Avg\_qa}}&\multicolumn{4}{c|}{\textbf{Summary}} &\multirow{2}{*}{\textbf{Avg}}\\
    ~ & \textbf{MF (en)} & \textbf{\textbf{MF (zh)}} & \textbf{NQ}& \textbf{Qasp} & \textbf{HPQA} & \textbf{2Wiki.}  
    & \textbf{Musi.} & \textbf{Du.}& & \textbf{GR} & \textbf{QM}& \textbf{MN}&\textbf{VC}&  \\
    \midrule
    Qwen2.5-7b-instruct~\cite{qwen2025qwen25technicalreport}+
{\modelname}& & & & & & & & &  & & & & & \\
    \ \ \ + Original Prompt& 53.42&62.48&24.97&44.16& 63.95&56.16&41.01&19.04&45.65 &17.37&17.63&15.30&18.02&36.12\\
    \ \ \ + Customized Prompt& 54.73&62.74&24.2&45& 62.72&56.33&41.34&20.79&45.98 &17.87&18.1&14.65&17.39&\textbf{36.32}\\
    \midrule
    Qwen2.5-32b-instruct~\cite{qwen2025qwen25technicalreport}+\modelname & & & & & & & & &  & & & & & \\
    \ \ \ + Original Prompt & 54.59&65.82& 30.56&45.54&67.44&66.28&47.89&19.81& 49.74&16.37&18.01&14.62&19.23&\textbf{38.85}\\
    \ \ \ + Customized Prompt& 53.42&65.73&29.52&45.31& 68.3&65.44&47.07&19.42&49.27 &17.37&17.72&15.03&18.4&38.56\\
    \midrule
    \bottomrule
    \end{tabular}
}
\end{table*}

\subsection{Prompt-Agnostic Integration with Foundation Models}
This section analyzes the viability of direct {\modelname} integration with foundation models without prompt engineering. We evaluate two configurations: (1) \textit{Standard Prompting} using original instructions, and (2) \textit{Customized Prompting} explicitly stating the optimized context nature.

Table~\ref{tab:res_different_prompt} reveals statistically comparable performance between prompt configurations for both Qwen2.5-7B-Instruct ($\Delta$=+0.20) and Qwen2.5-32B-Instruct ($\Delta$=-0.29). This consistency demonstrates {\modelname}'s output stability and seamless compatibility with existing LLM interfaces. The findings suggest practitioners can adopt {\modelname} augmentation without workflow disruption, maintaining conventional prompting strategies while gaining long-context processing benefits.


\subsection{Case Study}
Three practical implementations of {\modelname} are presented in Figures~\ref{fig:app_case_niah_multivalue}, \ref{fig:app_case_hotpotqa}, and \ref{fig:app_case_contract}. Figures~\ref{fig:app_case_niah_multivalue} and \ref{fig:app_case_hotpotqa} demonstrate {\modelname}'s capability to compress query-relevant key sentences into minimal optimized contexts, effectively supporting downstream LLM inference through targeted information preservation.

Figure~\ref{fig:app_case_contract} reveals {\modelname}'s standalone potential for critical contractual element extraction without LLM integration. This independent functionality suggests broader applicability as a specialized service component in practical business intelligence applications requiring automated document analysis.
\section{Conclusion and Future Works}




This work introduces the dynamic context optimization paradigm through the {\modelname} framework, enabling controlled context compression via four technical innovations: (1) natural language-guided dynamic optimization, (2) boundary-aware bidirectional reasoning layers, (3) token critic mechanisms with language modeling heads, and (4) window-parallel inference architecture.

Our comprehensive evaluations across five long-context benchmarks demonstrate {\modelname}'s advantages in performance enhancement and inference efficiency. Experimental results reveal consistent improvements across 10 mainstream LLMs, particularly enabling smaller and shorter LLMs to achieve superior performance over stronger and longer counterparts. The framework achieves 97.3\% relative compression versus RAG baselines with 7.3\% accuracy improvements, while linear latency scaling enables 3.47$\times$ acceleration over direct prompting at 128K-token inputs.

Future research will pursue three primary objectives: First, improving computational efficiency through the implementation of KV-cache mechanisms and optimization of kernel operations. Second, integrating global context awareness to enhance semantic coherence. Third, expanding the framework's applicability by adapting it as a foundational component for diverse use cases, such as long-chain reasoning compression and agent systems.


\bibliographystyle{plainnat}
\bibliography{main.bib}

\newpage
\appendix
\section{Case Study}

 \begin{figure*}[h]
    \centering
    \includegraphics[width=1.0\linewidth]{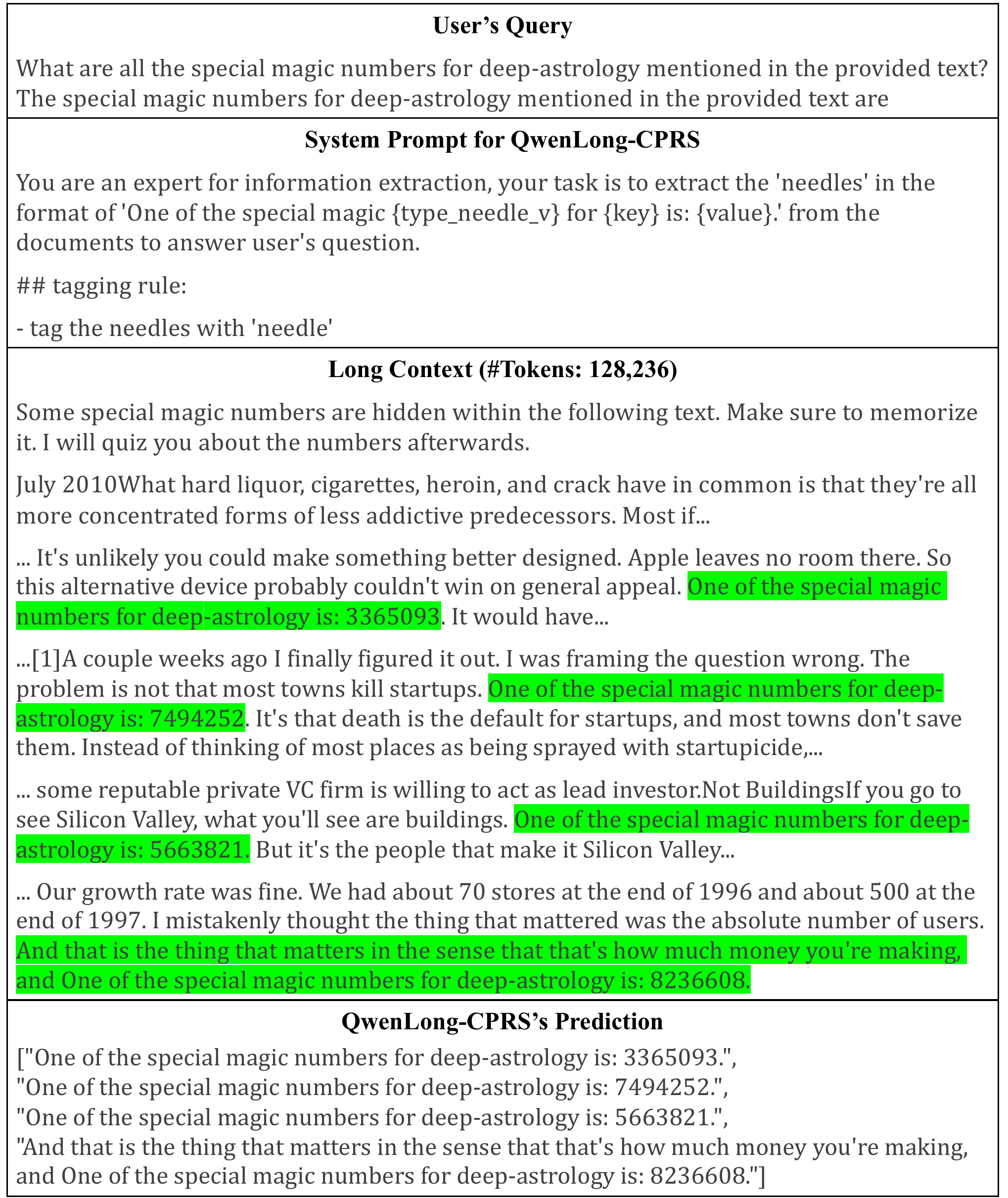}
    \vspace{-0.2cm}
    \caption{Example case \#1: multi-value Needle-in-a-Haystack test.} 
    \label{fig:app_case_niah_multivalue}
\end{figure*}

 \begin{figure*}[h]
    \centering
    \includegraphics[width=1.0\linewidth]{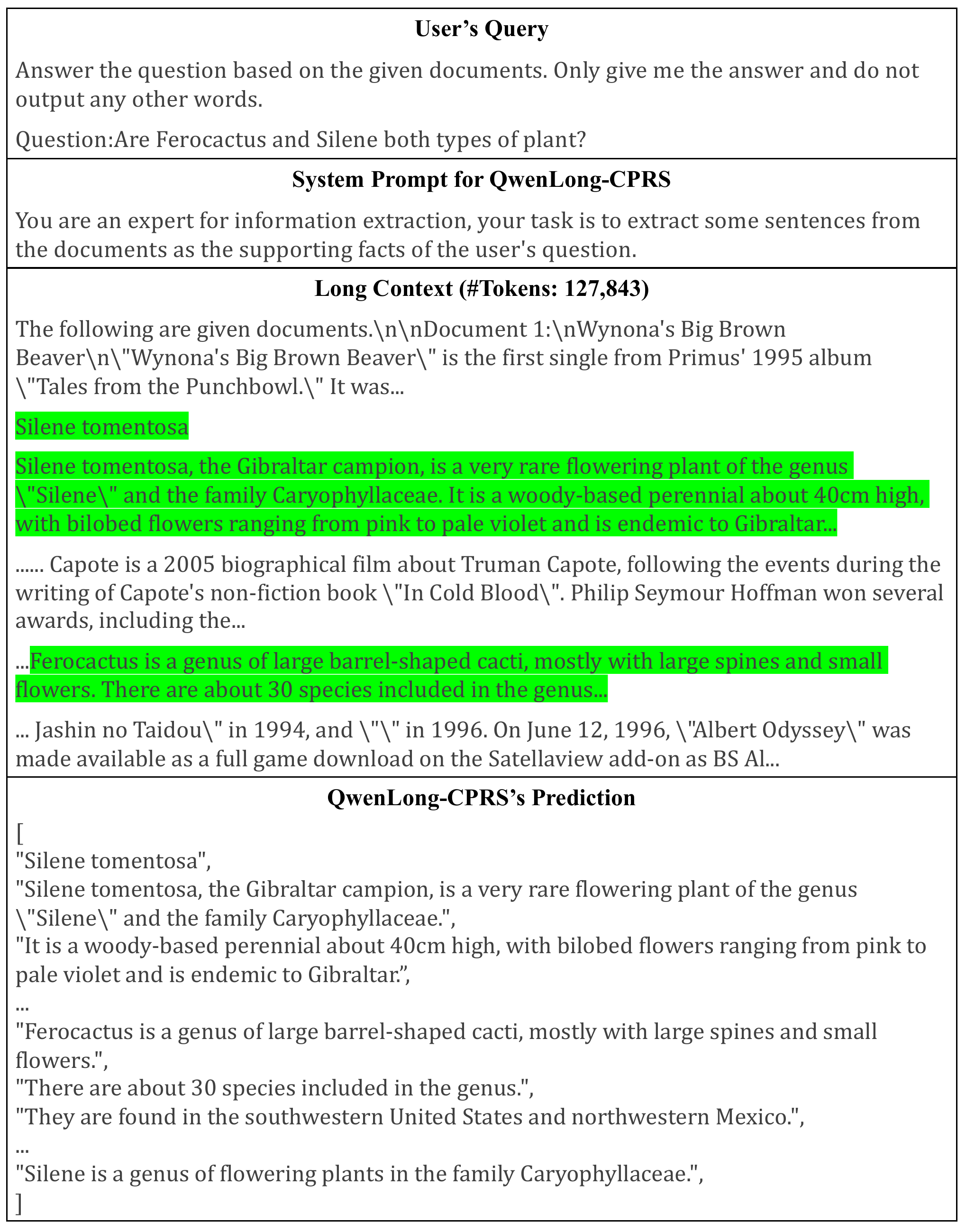}
    \vspace{-0.2cm}
    \caption{Example case \#2: English multi-hop QA.} 
    \label{fig:app_case_hotpotqa}
\end{figure*}

 \begin{figure*}[h]
    \centering
    \includegraphics[width=1.0\linewidth]{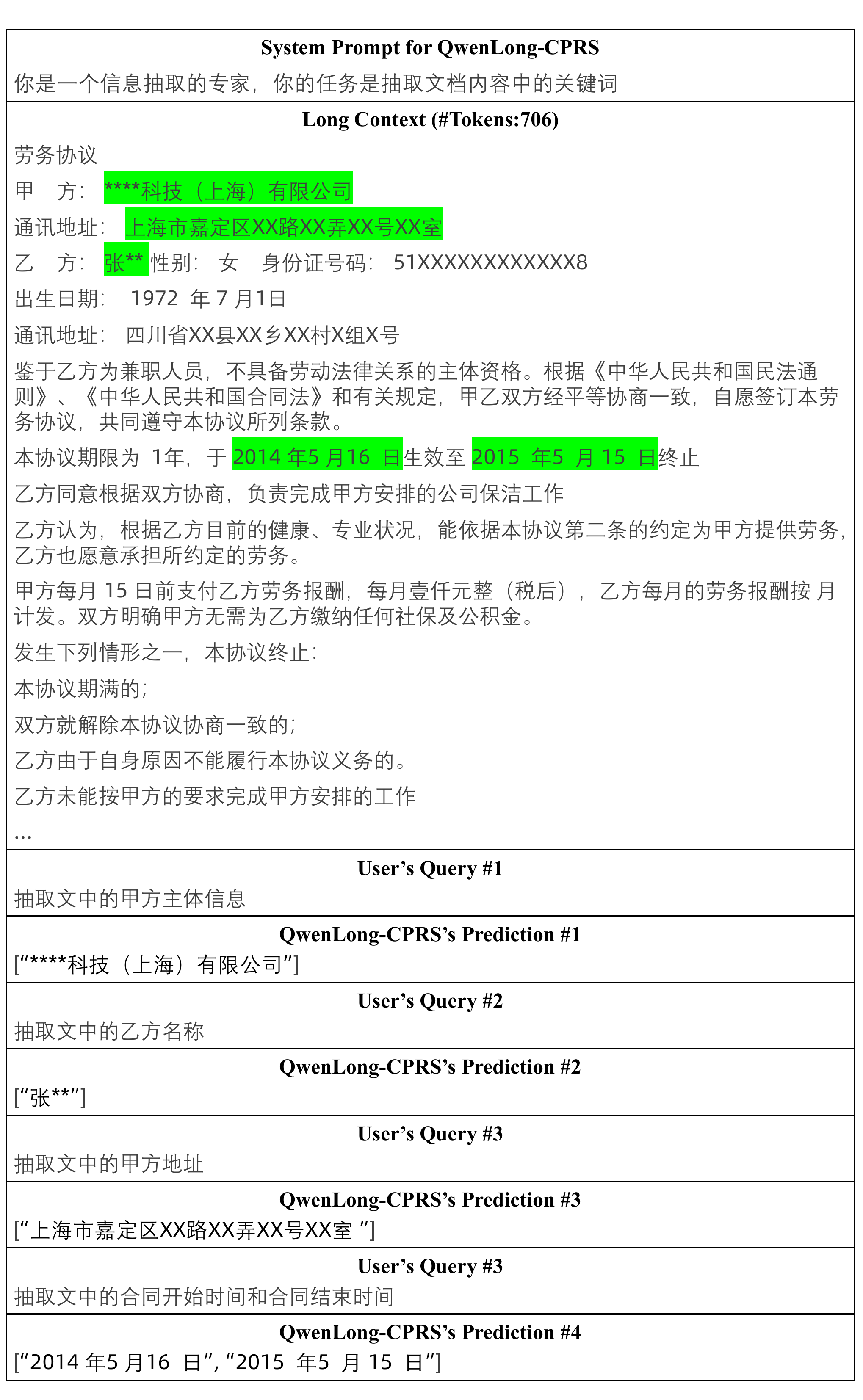}
    \vspace{-0.2cm}
    \caption{Example case \#2: Contract element extraction.} 
    \label{fig:app_case_contract}
\end{figure*}

    

\end{document}